%% file: 0_main.tex
\documentclass[sigconf,nonacm]{acmart}

\usepackage{booktabs}
\usepackage{graphicx}
\usepackage{amsmath}
\usepackage{amsfonts}
\usepackage{multirow}
\usepackage{color}
\usepackage{soul}
\usepackage{array}

\makeatletter
\let\acm@nolinkurl\nolinkurl
\renewcommand{\nolinkurl}[1]{\href{mailto:#1}{\acm@nolinkurl{#1}}}
\makeatother

\begin{document}
\title{A100: A Hybrid Vision Foundation Model for Agricultural Plant Image Analysis}

\author{Mayur Pratim Das}
\affiliation{\institution{Annam.AI, IIT Ropar}\city{Rupnagar}\country{India}}
\email{2023csm1007@iitrpr.ac.in}

\author{Ajai Dubey}
\affiliation{\institution{IIT Ropar}\city{Rupnagar}\country{India}}
\email{ajai.25aiz0023@iitrpr.ac.in}

\author{Aksshay Mathew P}
\affiliation{\institution{IIT Ropar}\city{Rupnagar}\country{India}}
\email{2024aim1013@iitrpr.ac.in}

\author{Sana Afreen}
\affiliation{\institution{Annam.AI, IIT Ropar}\city{Rupnagar}\country{India}}
\email{sanaafreen@annam.ai}

\author{Rishi Kumar Soni}
\affiliation{\institution{Annam.AI, IIT Ropar}\city{Rupnagar}\country{India}}
\email{staff.rishi.soni@iitrpr.ac.in}

\author{Deepak Meena}
\affiliation{\institution{Annam.AI, IIT Ropar}\city{Rupnagar}\country{India}}
\email{staff.deepak.meena@iitrpr.ac.in}

\author{Pratibha Kumari}
\affiliation{\institution{Dept. of DSE, IISER Bhopal}\city{Bhopal}\country{India}}
\email{pratibha@iiserb.ac.in}

\author{Mukesh Saini}
\affiliation{\institution{Dept. of CSE, IIT Ropar}\city{Rupnagar}\country{India}}
\email{mukesh@iitrpr.ac.in}

\renewcommand{\shortauthors}{M. P. Das et al.}

\begin{abstract}
The use of accurate vision models for crop monitoring, disease diagnosis, and plant phenotyping in agriculture is hindered by the lack of expertly annotated data, especially in developing countries. We propose \textbf{A100}, an agricultural vision foundation model that is pre-trained using self-supervised learning on \textbf{3 million unlabeled images from fields} covering more than 100 crop species. The architecture of A100 consists of hybrid CNN-Transformer backbone, which involves the fusion of MBConv stages for local feature extraction and Transformer stages for global contextual representation. A100 uses pre-training with the integration of DINO self-distillation, iBOT masked patch reconstruction, and KoLeo uniformity regularization, resulting in features that are better than ImageNet-12K supervised representations for agricultural applications without any fine-tuning. By utilizing a frozen backbone, the accuracy for crop classification of A100 is \textbf{95.37\%}, which is \textbf{+2.56\%} better than ImageNet-12K, and reduces the leaf count error by \textbf{10.4\%} on the GroMo25~\cite{gromo2025} phenotyping benchmark, proving that the same backbone can be used for classification and regression tasks. After full fine-tuning, the performance of A100 is \textbf{99.90\%} on 32-class crop identification and \textbf{99.02\%} on 44-class disease detection. On the difficult PlantDoc dataset, A100 outperforms all baselines, such as CvT-13 (64.89\% against 63.56\%). Remarkably, A100 obtains \textbf{98.01\%} accuracy on plant classification using just 1\% of the labelled data (about 284 images per category), showing great label efficiency for use cases with limited labeling. Under an identical protocol, A100 also outperforms the publicly released agricultural foundation models FoMo4Wheat and LeafVision by more than \textbf{14 points} at 1\% labels. We plan to publicly release our pretrained model to allow agricultural professionals to adapt A100 for various plant analysis applications using limited annotations.
\end{abstract}



\maketitle

\input{1_intro_and_literature}

\input{2_method}

\input{3_results}

\section{Conclusion}\label{conclusion}
In this paper, we have proposed A100, a self-supervised hybrid CNN-Transformer foundation model for agricultural plant image analysis. By using multi-objective self-supervised learning (SADG) to pre-train on ~3 million unlabeled images across 100 crop species, A100 provides a robust initialization for a wide range of agricultural vision tasks while avoiding the need for labeled data during pre-training.  A100 achieves 99.90\% accuracy on crop classification and 99.02\% on disease detection with 73M parameters, which is comparable or superior to an ImageNet-12K supervised baseline on all tasks.

The most important contribution of A100 is the quality of the frozen backbone's representation, as linear-probe evaluation on crop identification (+2.56\%) and disease detection (+0.97\%) without fine-tuning shows it is superior. On external benchmarks, without any data overlap, A100 improves the performance of all baseline models on PlantDoc (+4.30\% over ImageNet-12K) and PlantWild (+1.21\%), with 99.96\% on PlantVillage. Furthermore, A100 generalizes beyond classification to regression-based plant phenotyping, reducing leaf counting error by 10.4\% over ImageNet on the GroMo25~\cite{gromo2025} benchmark. Compared directly with publicly released agricultural foundation models under the same protocol, A100 leads the strongest of them, FoMo4Wheat \cite{han2025fomo4wheat}, by 14.2 points with only 1\% of the labels, showing that a broad multi-crop field corpus yields representations that narrower agricultural pretraining does not.

We plan to publicly release the pretrained A100 weights to enable the agricultural research community to fine-tune the model for crop identification, disease diagnosis, pest surveillance, and related tasks, even with limited annotated data. Future work will extend A100 to multi-label settings, additional crop types, and integration with vision-language capabilities for zero-shot agricultural diagnosis which will be crucial to real-world agricultural challenges.  

\bibliographystyle{ACM-Reference-Format}
\bibliography{5_reference}

\end{document}

%% file: 1_intro_and_literature.tex
\begin{figure}[!t]
    \centering
                \includegraphics[width=1\linewidth]{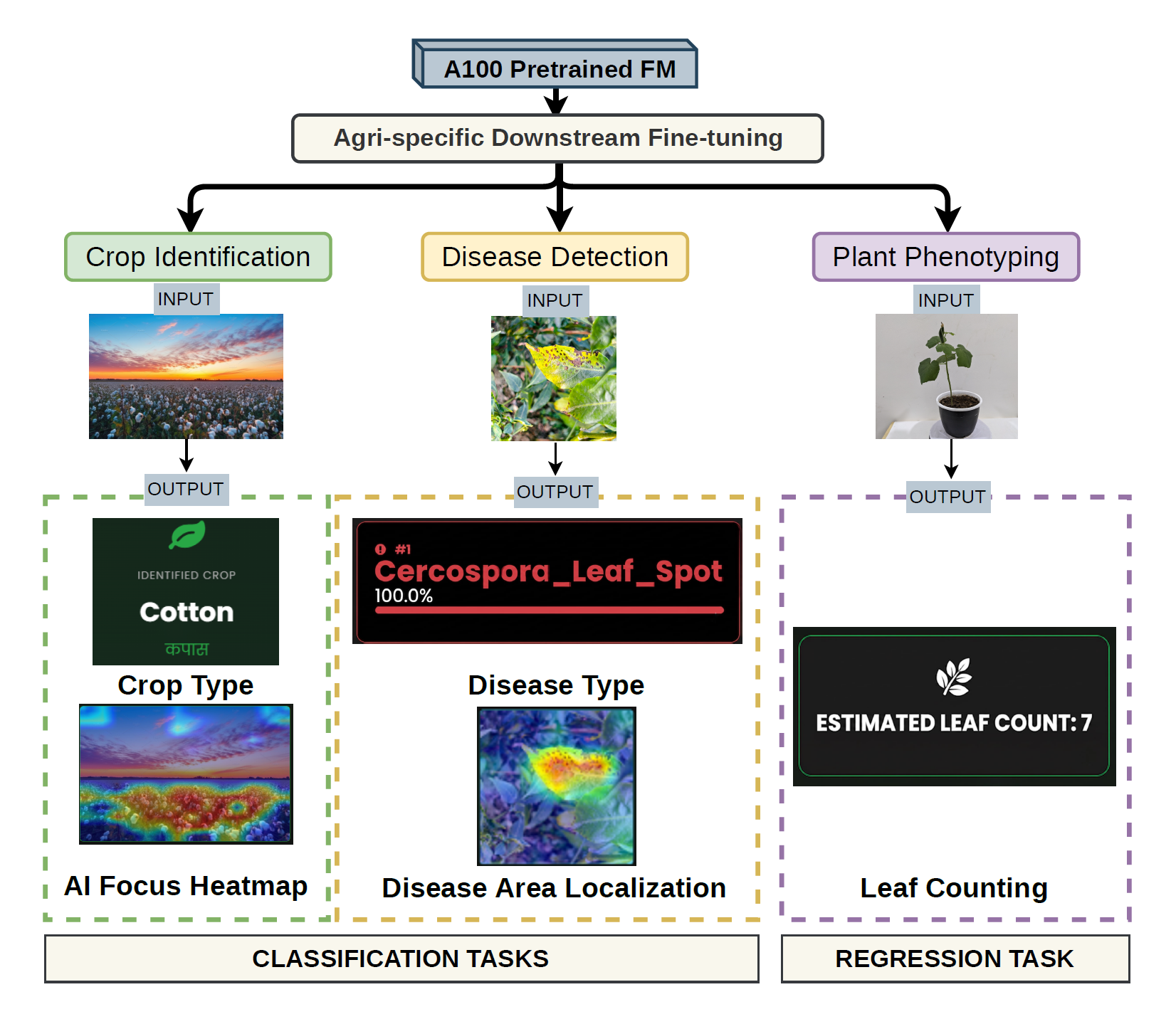}
    \caption{Deployment on downstream tasks for Crop Identification, Disease Detection, and Plant Phenotyping (Leaf Counting)}
    \label{fig:intro_gradcam}
\end{figure}
\section{Introduction}
Plant image analysis is crucial for crop surveillance, in agricultural automation, and farmer decision support systems. However, there is a scarcity of good-quality labeled data.       
Foundation Models (FMs) pretrained on large amounts of data via Self-Supervised Learning (SSL) \cite{bommasani2021opportunities} have emerged as a paradigm shift for overcoming the scarcity of annotations. DINO \cite{caron2021emerging}, DINOv2 \cite{oquab2024dinov2}, and MAE \cite{he2022masked} learn powerful, transferable representations without human labels. This principle is compelling for agriculture, where annotated datasets are limited. Domain-specific FMs have demonstrated substantial gains over generic ImageNet-pretrained counterparts in medicine \cite{medsam}, underwater and marine imagery \cite{aquaticclip}, and remote sensing \cite{hong2024spectralgpt}, motivating domain-specific pretraining for agriculture.

Although there are FMs that are highly optimized for remotely sensed agriculture images \cite{luo2025phenoyieldnet,chiu2020agriculturevision}, they struggle to capture the fine-grained details of proximally captured images (e.g., smartphone images) needed for tasks such as identifying crops and detecting plant diseases. 
The most closely related work, Agri-FM+ \cite{alnahian2025agrifm}, demonstrates that continual self-supervised adaptation of a ResNet-50 backbone on 147K agricultural images yields improvements over an ImageNet initialization. 
 The pretraining corpus of Agri-FM+ (147K images) is small, limiting representational diversity. It uses a pure CNN backbone (ResNet-50), which lacks mechanisms for global context modeling critical to understanding whole-plant morphology.

\begin{figure*}[!t]
\centering
\includegraphics[scale=0.45]{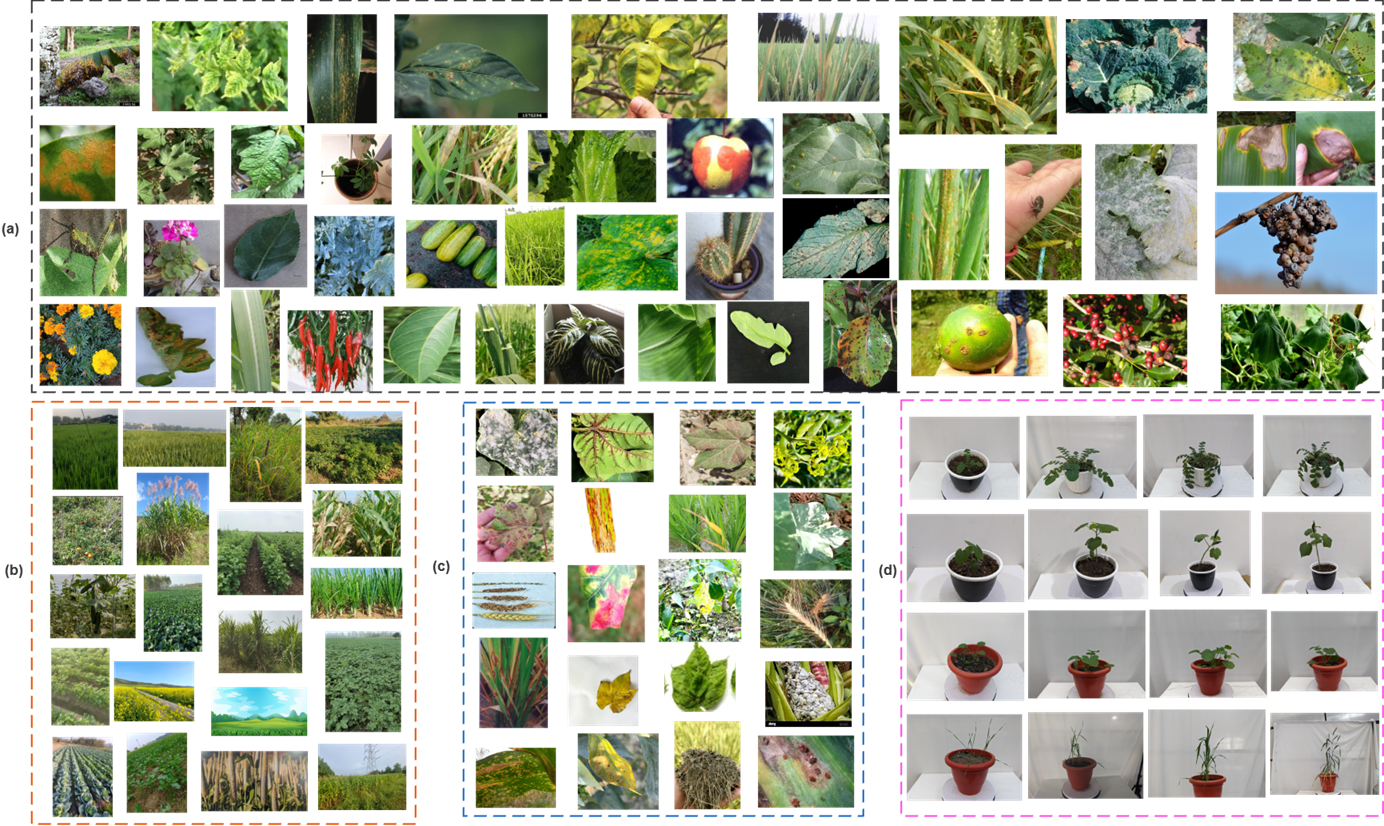}
\caption{Representative samples from the datasets used for (a) self-supervised pretraining, (b) crop classification, (c) disease detection, and (d) plant phenotyping.}
\label{fig:dataset_sample}
\end{figure*}

To address these limitations, we propose A100, a hybrid CNN-Transformer vision FM designed specifically for plant image analysis (see Fig.~\ref{fig:intro_gradcam}). A100 employs a hybrid backbone combining MBConv stages for local texture extraction with Transformer stages for global context, pretrained via Stage-Aware Dual-Granularity (SADG) self-supervised learning on 3 million unlabeled agricultural images spanning over 100 crop species. 
When evaluated on three downstream agricultural tasks, A100 achieves 99.90\% accuracy for 32-class crop classification, 99.02\% for 44-class disease detection, and reduces leaf counting error by 10.4\% over ImageNet on the GroMo25 plant phenotyping benchmark. The main contributions of this work are summarized as follows:

\begin{itemize}
\item We curate and pretrain on \textbf{3~million unlabeled agricultural field images} spanning 100~crop species- approximately $20\times$ larger than the prior domain-specific corpus Agri-FM+ (147K).

\item We propose a multi-objective self-supervised pretraining framework with DINO self-distillation, iBOT masked patch reconstruction, and KoLeo uniformity regularization to generate representations that are \textbf{2.56\% better than ImageNet-12K supervised features} under linear probe on crop identification and \textbf{4.30\% better than the identical ImageNet-12K-initialized model} on the challenging PlantDoc benchmark.

\item We demonstrate that A100 achieves \textbf{98.01\% crop classification accuracy with only 1\% of labeled data}, enabling deployment in data-scarce agricultural settings. We plan to publicly release the pretrained weights of A100 for the research community.

\item We benchmark A100 against publicly released agricultural foundation models under an identical protocol: with 1\% of the labels, A100 \textbf{outperforms FoMo4Wheat\cite{han2025fomo4wheat} and LeafVision \cite{han2026leafvision} by 14.2--20.2 points}, consistently exhibiting sub-point variance across random seeds.

\end{itemize}
The remainder of this paper is organized as follows. Section 2 reviews related work, Section 3 presents the proposed methodology, Section 4 reports and discusses the experimental results, and Section 5 concludes the paper.


\begin{table*}[!ht]
\centering
\footnotesize
\setlength{\tabcolsep}{3pt}
\caption{Composition of the pretraining corpus and downstream fine-tuning datasets.}
\label{tab:all_datasets}
\resizebox{\textwidth}{!}{
\begin{tabular}{l r l r l r}
\toprule
\multicolumn{2}{c}{\textbf{Pretraining Corpus}} &
\multicolumn{2}{c}{\textbf{Crop Identification}} &
\multicolumn{2}{c}{\textbf{Disease Detection}} \\
\multicolumn{2}{c}{\textit{(${\sim}$3,000,000 Images, 100 Crops)}} &
\multicolumn{2}{c}{\textit{(1,142,326 Images, 32 Classes)}} &
\multicolumn{2}{c}{\textit{(51,994 Images, 44 Classes)}} \\
\cmidrule(lr){1-2}\cmidrule(lr){3-4}\cmidrule(lr){5-6}
\textbf{Source} & \textbf{Images} & \textbf{Crop Class} & \textbf{Images} & \textbf{Crop--Disease} & \textbf{Images} \\
\midrule
\multicolumn{2}{l}{\textit{Proprietary On-Field Collection}} & & & & \\
~~Field agents (100 crops, multi-state)  & 1{,}367{,}927 & Wheat        & 123{,}631 & Chilli--Leaf Curl     & 2{,}608 \\
~~Farm survey imagery (diverse fields)   & 110{,}538 & Sugarcane    & 84{,}127  & Cotton--Healthy       & 4{,}869 \\
                                         &           & Gram         & 77{,}779  & Rice--Brown Spot      & 3{,}721 \\
\cmidrule(lr){1-2}
\multicolumn{2}{l}{\textit{Public Crop \& Weed Datasets}} & & & & \\
~~CWD30~\cite{ilyas2025cwd30} (30 crop-weed classes)       & 219{,}770 & Rice         & 65{,}942  & Rice--Blast           & 3{,}022 \\
~~CCMT~\cite{mensah2023ccmt} (crop classification)           & 152{,}738 & Mustard      & 61{,}084  & Cotton--Vertic.\ Wilt & 2{,}031 \\
~~Agriculture Vision~\cite{chiu2020agriculturevision} (aerial)    & 94{,}986  & Maize        & 46{,}686  & Wheat--Wheat Blast    & 1{,}382 \\
~~Web-scraped agricultural imagery                 & 699{,}062 & Bajra        & 33{,}832  & Maize--Rust Leaf      & 1{,}488 \\
~~DeepWeeds~\cite{alex2017deepweeds} (9 species)           & 17{,}509  & Cotton       & 28{,}237  & Rice--Tungro          & 3{,}260 \\
\cmidrule(lr){1-2}
\multicolumn{2}{l}{\textit{Public Disease Datasets}} & & & & \\
~~IP102 Insect~\cite{wu2019ip102} (102 classes)          & 75{,}222  & Onion        & 29{,}288  & Maize--Gray L.\ Spot  & 330 \\
~~Multi-crop disease compilation                   & 79{,}915  & Brinjal      & 59{,}446  & Cotton--Boll Rot      & 950 \\
~~TOM2024 (tomato diseases)~\cite{appiah2024tom2024}                        & 25{,}844  & Chilli       & 26{,}861  & Rice--Downy Mildew    & 730 \\
~~Plum Leaf/Fruit diseases~\cite{nayeem2025plum}                         & 21{,}554  & Pea          & 24{,}948  & Wheat--Mildew         & 1{,}010 \\
~~Maize Leaf Disease~\cite{wiesnerhanks2018image}                               & 18{,}222  & Potato       & 11{,}452  & Maize--Leaf Blight    & 1{,}205 \\
~~Wheat Plant Disease~\cite{kushagra2024wheat}                              & 14{,}155  & Cabbage      & 18{,}694  & Rice--Sheath Blight   & 890 \\
~~Sugarcane, Rice Leaf, Dragon Fruit, others   & 32{,}881  & Cucumber     & 19{,}728  & Chilli--Anthracnose   & 1{,}150 \\
\cmidrule(lr){1-2}
\multicolumn{2}{l}{\textit{Growth \& Phenology}} & & & & \\
~~Crop growth stages                               & 121{,}302 & Tomato       & 37{,}615  & Chilli--Bact.\ Spot   & 1{,}980 \\
~~PhenoBench (crop phenotyping)                    & 13{,}761  & Radish       & 25{,}289  & Cotton--Curl Virus    & 690 \\
~~Rice varieties                                   & 9{,}881   & Cauliflower  & 18{,}667  & Wheat--Smut           & 1{,}000 \\
~~Wheat damage assessment                          & 4{,}524   & + 14 more    & 349{,}020 & + 26 more classes     & 19{,}678 \\
\midrule
\textbf{Total (deduplicated)}      & \textbf{${\sim}$3{,}000{,}000} & & \textbf{1{,}142{,}326} & & \textbf{51{,}994} \\
\bottomrule
\end{tabular}
}
\vspace{1mm}
\end{table*}

\section{Related Work}
This section reviews prior work on SSL strategies, existing vision FMs, and hybrid CNN-Transformer architectures. 

\subsection{Self-Supervised Learning for Plant Image Analysis}
SSL reduces dependence on annotated data by learning representations directly from unlabeled images \cite{caron2021emerging, oquab2024dinov2}. In agriculture, contrastive SSL methods like PlantCLR \cite{shah2026plantclr}, ConMamba \cite{mamun2025conmamba}, and VICReg-based models \cite{safonova2026self} demonstrate strong performance for disease and crop classification, even under 5\% label scarcity \cite{safonova2026self}. Most closely related, LeafVision \cite{han2026leafvision} systematically evaluates SimCLR \cite{chen2020simclr}, BYOL \cite{grill2020byol}, and DINO \cite{caron2021emerging} across six backbones on 540K leaf images, establishing DINO's superiority. However, existing agricultural SSL models evaluate individual backbones in isolation and target single-task classification. In contrast, A100 pairs DINO self-distillation with a 3M-image corpus and a unified hybrid architecture, evaluated across three diverse tasks.

\subsection{Agricultural Vision Foundation Models}
Recent research has focused on adapting vision FMs for precision agriculture \cite{hong2024spectralgpt}. Domain-specific vision FMs address ImageNet's domain shift in agricultural settings \cite{bommasani2021opportunities}. Early FMs include Agri-FM+ \cite{alnahian2025agrifm} and FoMo4Wheat \cite{han2025fomo4wheat}. FoMo4Wheat is pretrained on 2.5M wheat images and targets wheat canopy- and organ-level tasks, whereas Agri-FM+ reports only detection and segmentation results and has not publicly released its pretrained weights. Recent multi-modal vision-language models (VLMs) such as SCOLD \cite{nguyen2026scold} (186K image-captions), AgriCLIP \cite{agriclip}, and AgriRAG \cite{marques2026agrirag} enable zero-shot diagnosis and diagnostic reasoning, while PhenoYieldNet \cite{luo2025phenoyieldnet} applies temporal SSL for crop yield prediction. Nevertheless, multi-modal FMs require textual descriptions rarely available in the field, whereas vision-only FMs \cite{alnahian2025agrifm} rely on monolithic backbones. A100 is a vision-only FM operating on RGB images alone, leveraging a hybrid CNN-Transformer architecture pretrained on 3M images.

\subsection{Hybrid CNN-Transformer Architectures}
Hybrid CNN-Transformer models combine CNN local feature extraction with Transformer global self-attention \cite{dai2021coatnet, guo2022cmt}. In agricultural image analysis, hybrid frameworks \cite{appiah2025hybrid, deepika2025mobileplantvit, jawed2026hybrid} concatenate CNN features with Vision Transformers for foliar disease classification. Notably, LSeTNet \cite{ahmmed2026lsetnet} integrates Squeeze-and-Excitation (SE) attention with transformers for leaf disease detection, while HyPest-Net \cite{sharma2025hypestnet} combines CNN spatial/channel attention with ViT-B/16 for pest identification \cite{suzauddola2026advanced, selinchandra2025comparative}. Crucially, all existing agricultural hybrid models are fully supervised, task-specific classifiers trained on small datasets. A100 bridges this gap by introducing the first self-supervised hybrid CNN-Transformer foundation model for agricultural vision, pretrained via multi-objective SADG on 3M images and publicly released for the research community.

%% file: 2_method.tex
\begin{figure*}[t]
    \centering
    \includegraphics[width=\linewidth]{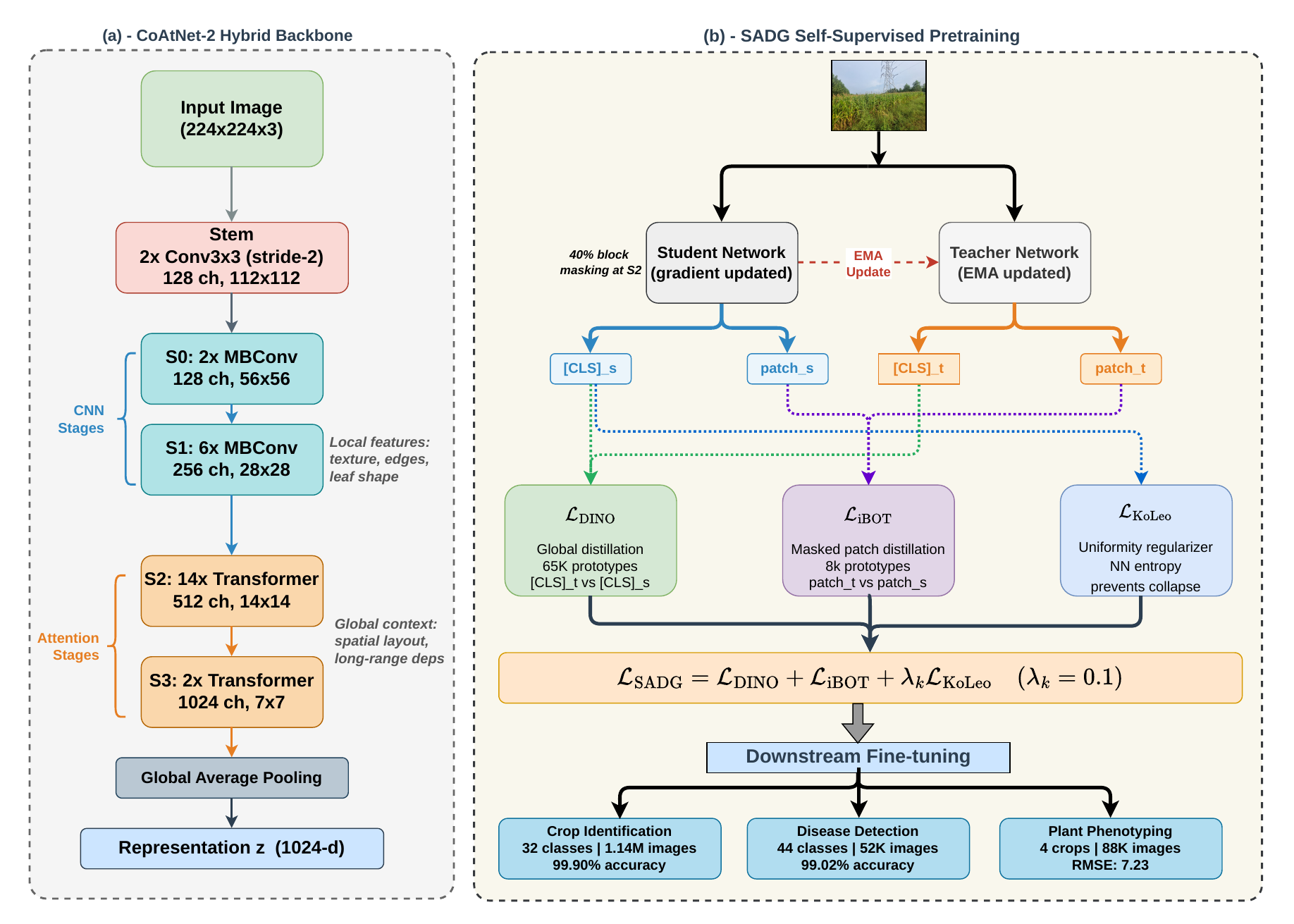}
    \caption{A100 architecture overview: (a) CNN-Transformer hybrid backbone with MBConv stages (S0--S1) and Transformer stages (S2--S3), (b) SADG pretraining framework with three self-supervised objectives.}
    \Description{A block diagram showing the A100 architecture. The left panel shows the CNN-Transformer hybrid backbone processing an input image through stem, MBConv, and Transformer stages. The right panel shows the SADG self-supervised pretraining framework utilizing Student and Teacher networks with DINO, iBOT, and KoLeo objectives.}
    \label{fig:architecture}
\end{figure*}

\section{Methodology}\label{gen_inst}

In this section, we present a hybrid vision foundation model, A100, for analyzing proximally sensed plant images. The model backbone is based on a hybrid CNN-transformer integrated architecture \cite{dai2021coatnet} and it is trained via multi-objective self-supervised pretraining (SADG). The datasets, problem formulation, model architecture, SADG pretraining framework, and task-specific fine-tuning are described in the following subsections.

\subsection{Datasets and Data Preparation} 

We constructed a large-scale unlabeled agricultural image corpus comprising approximately 3M images collected from 17 different sources covering 100+ crop species, disease symptoms, growth stages and agricultural field conditions. Approximately half of the corpus ($\approx$1.5 million) consists of proprietary in-field images collected by our agricultural field agents and farm surveys across multiple farming regions in India, while the remaining images were aggregated from diverse public crop, weed, disease, pest, and phenotyping datasets. This combination provides broad coverage of both controlled and real-world agricultural conditions for representation pretraining. An exhaustive list of the dataset sources used to construct the pretraining corpus, along with the corresponding number of images from each source, is provided in Table~\ref{tab:all_datasets}.

After aggregation, approximately 90K duplicate images were removed using file name and file size-based heuristics, resulting in a deduplicated corpus of approximately 3M images. For efficient distributed pretraining, the images were organized into 380 WebDataset~\cite{webdataset} TAR shards, with each shard containing approximately 8K images. This organization enables sequential I/O and reduces data-loading overhead during large-scale training.

For downstream evaluation, we consider three agricultural vision tasks, whose compositions are also reported in Table~\ref{tab:all_datasets}. The crop identification dataset contains 1,142,326 field images distributed across 32 crop classes. The disease detection dataset consists of 51,994 images of five main crops, namely Chili, Cotton, Maize, Rice, and Wheat, with 44 classes of plant disease. For the plant phenotyping, we used the GroMo25 dataset~\cite{gromo2025}, which is part of (ACM Multimedia 2025) and contains ${\sim}$88K multiview images of four crops (Wheat, Okra, Radish, Mustard), captured under controlled conditions across multiple growth stages with per-image leaf count annotations (range: 2--40 leaves). Further, three external benchmark datasets including PlantDoc~\cite{singh2020plantdoc}, PlantVillage~\cite{hughes2015open}, and PlantWild~\cite{wei2024benchmarking} are considered for disease detection downstream evaluation. Each downstream dataset is divided into training, validation, and test sets.

We evaluated crop identification and disease detection using a stratified per-class random split (80/10/10, fixed seed 42). We applied this split once at the image level before training and kept it strictly consistent across all methods: 913,848 / 114,218 / 114,260 crop images and 41,595 / 5,199 / 5,200 disease images (train/val/test). We drew our low-label subsets (1\% and 10\%) exclusively from the training split, keeping the full validation and test sets intact. Furthermore, we adopted the official challenge split for GroMo25 and standard published splits for external benchmarks. To strictly prevent data leakage, we excluded all downstream evaluation images from our pretraining corpus. We show representative samples from both phases in Fig.~\ref{fig:dataset_sample}.

\subsection{Problem Formulation}
We tackle visual recognition in a two-stage setting, in which the model is first pretrained using self-supervised learning, and then fine-tuned for the specific task.

\subsubsection{Self-supervised pretraining}
Given an unlabeled corpus $\mathcal{D}_u = \{\mathbf{x}_i\}_{i=1}^{N_u}$ of $N_u \approx 3\text{M}$ images, learn a feature encoder $f_\theta: \mathbb{R}^{3 \times H \times W} \to \mathbb{R}^{D}$ (where $D = 1024$) via SADG multi-objective self-supervised learning, producing representations informative for downstream tasks without using any labels.

\subsubsection{Fine-Tuning}
For crop identification on a labeled dataset $\mathcal{D}_{\text{crop}} = \{(\mathbf{x}_i, y_i)\}_{i=1}^{N_{\text{crop}}}$ where $y_i \in \{1, \ldots, 32\}$, learn a classifier $g_{\phi}: \mathbb{R}^{D} \to \Delta^{31}$, which maps the pretrained representation $f_\theta(\mathbf{x})$ of the image $\mathbf{x}$ to the number $y$ denoting the crop species. Similarly for disease detection ($\mathcal{Y}_{\text{disease}} = \{1, \ldots, 44\}$). For leaf counting, a regression head maps $f_\theta(\mathbf{x})$ to a continuous scalar predicting the number of leaves.

\subsection{A100 Architecture}\label{sec:architecture}

A100 employs a hybrid model that integrates convolutional and attention mechanisms in a unified architecture (see Fig.~\ref{fig:architecture}) \cite{dai2021coatnet}. The details of the architecture are given below.

\subsubsection{Hybrid Design}
Images of agricultural crops contain visual patterns at various scales: fine-grained patterns (such as disease lesions and leaf morphology) can be processed with local features, whereas global reasoning is needed to distinguish similar patterns or to understand the context of the whole crop. The hybrid  model tackles both of these in its staged design:

\begin{itemize}
    \item \textbf{Stages S0--S1 (MBConv):} Mobile inverted bottleneck convolution blocks are used to extract local features, which are translation equivariant. They are the key stages for picking up fine texture, which is essential for spotting disease spots, leaf margins, and leaf venation patterns.
    \item \textbf{Stages S2--S3 (Transformer):} Long-range dependencies between spatial features are captured by relative self-attention blocks. These stages allow the model to incorporate information from the entire image, which is necessary to differentiate diseases that look alike at the local scale but differ in spatial distribution.
\end{itemize}

\subsubsection{Architecture Configuration}
The backbone processes $224 \times 224$ input images through:
\begin{itemize}
    \item \textbf{Stem:} Two stride-2 convolutions ($3 \to 128$ channels, $112 \times 112$)
    \item \textbf{S0:} 2 MBConv blocks ($128$ ch, $56 \times 56$)
    \item \textbf{S1:} 6 MBConv blocks ($256$ ch, $28 \times 28$)
    \item \textbf{S2:} 14 Transformer blocks ($512$ ch, $14 \times 14$)
    \item \textbf{S3:} 2 Transformer blocks ($1024$ ch, $7 \times 7$)
\end{itemize}
Global average pooling on S3 produces a 1024-dimensional representation. The entire backbone has 73M parameters. Initialization: We start from the ImageNet-12K pre-trained model and continually pre-train the model on our agricultural corpus using SADG.

\subsection{Stage-Aware Dual-Granularity (SADG) Pretraining}
\label{sec:sadg}
SADG employs complementary self-supervised goals at various granularities of features to learn multi-scale agricultural representations. The specific objectives (DINO, iBOT, KoLeo) are listed in the SSL literature, while the combination of a hybrid CNN-Transformer backbone and domain specific design considerations (agriculture adapted augmentations, full backbone learning rate for fast domain adaptation from ImageNet initialization, asymmetric teacher-student augmentation) forms our pretraining framework. The novel algorithmic idea is not the most important part, but the realization that this novelty, when applied to a domain-specific hybrid architecture at scale (3M images, 100 crops) yields features that measurably outperform generic ImageNet pretraining and previous agricultural FMs.

\subsubsection{DINO Self-Distillation (Global)}
Global image representations are learned using cross-view consistency in a teacher--student framework. The teacher (soft targets updated by EMA) gives multiple augmented crops to the student. Stable training is achieved by using DINOv2-style Sinkhorn-Knopp normalization~\cite{oquab2024dinov2}:

\begin{equation}
\label{eq:dino_loss}
\mathcal{L}_{\text{DINO}} = -\sum_{t,\,s \neq t} \sum_{d=1}^{K} P_t^{(d)} \log P_s^{(d)},
\end{equation}
where $P_t$ and $P_s$ are the teacher and student distributions obtained via temperature-scaled softmax over $K = 65{,}536$ prototypes. The teacher temperature warms from 0.04 to 0.07; student temperature is fixed at 0.1.

\subsubsection{iBOT Masked Patch Distillation (Spatial)}
To learn local spatial representations, following iBOT~\cite{ibot}, we apply block-wise masking (40\% ratio, $2 \times 2$ blocks) at the S2 feature map ($14 \times 14$). The student must reconstruct teacher patch representations at masked positions:
\begin{equation}
\mathcal{L}_{\text{iBOT}} = -\sum_{m \in \mathcal{M}} \sum_{d=1}^{K'} Q_t^{(d)}(m) \log Q_s^{(d)}(m),
\end{equation}
where $\mathcal{M}$ denotes masked positions, $K' = 8{,}192$ patch prototypes, and $Q_t$, $Q_s$ are Sinkhorn-Knopp normalized teacher and student patch distributions.

\subsubsection{KoLeo Uniformity Regularization (Global)}
To prevent feature collapse, the Kozachenko-Leonenko entropy estimator encourages uniform coverage of the representation space:
\begin{equation}
\mathcal{L}_{\text{KoLeo}} = -\frac{1}{n} \sum_{i=1}^{n} \log d_{\text{NN}}(\mathbf{z}_i),
\end{equation}
where $d_{\text{NN}}(\mathbf{z}_i)$ is the nearest-neighbor distance for sample $i$ on the unit hypersphere.

\subsubsection{Projection Heads}
The distributions $P$ and $Q$ in the DINO and iBOT objectives are produced by two separate projection heads (Table~\ref{tab:heads}).
For the global objective, the representation $\mathbf{z} = \mathrm{GAP}(S3) \in \mathbb{R}^{1024}$ (the same representation used downstream) is mapped by the DINO head $h_D$, a three-layer MLP (1024$\to$2048$\to$2048$\to$256, each hidden layer followed by BatchNorm and GELU), whose 256-d output is L2-normalized and passed through a weight-normalized linear layer without bias (256$\to$$K$) whose rows act as learnable prototypes:

\begin{equation}
\label{eq:dino_head}
P_s = \mathrm{softmax}\!\big(h_D(\mathbf{z}_s)/\tau_s\big), \qquad
P_t = \mathrm{SK}\!\big(h_D(\mathbf{z}_t)/\tau_t\big),
\end{equation}
For the patch objective, the iBOT head $h_I$ applies $1{\times}1$ convolutions across the $14{\times}14$ feature map. At masked positions $m$, the student predicts softmax logits $Q_s(m)$ to match the unmasked teacher distribution $Q_t(m)$, which is normalized via 3-iteration Sinkhorn-Knopp ($\mathrm{SK}$). The backbone and both independent heads are duplicated for the teacher network, which is updated strictly via an exponential moving average (EMA) of the student parameters. Additionally, KoLeo regularization is applied to the L2-normalized student representation $\mathbf{z}_s$. Both heads are discarded post-pretraining, retaining only the 73M-parameter backbone for all downstream evaluations.

\begin{table}[!htbp]
\centering
\footnotesize
\caption{SADG Projection Heads}
\label{tab:heads}
\setlength{\tabcolsep}{3pt}
\begin{tabular}{lcc}
\toprule
 & \textbf{DINO head $h_D$} & \textbf{iBOT head $h_I$} \\
\midrule
Input & GAP(S3), 1024-d & S2, $512{\times}14{\times}14$ \\
Applied & per image & per spatial position \\
MLP & 1024$\to$2048$\to$2048$\to$256 & 512$\to$2048$\to$2048$\to$256 \\
Prototypes & $K = 65{,}536$ & $K' = 8{,}192$ \\
Teacher norm. & Sinkhorn-Knopp & Sinkhorn-Knopp \\
Parameters & ${\approx}$23.6M & ${\approx}$7.9M \\
\bottomrule
\end{tabular}
\end{table}

\subsubsection{Combined SADG Objective}
The primary objectives are active from the beginning of training:
\begin{equation}
\mathcal{L}_{\text{SADG}} = \mathcal{L}_{\text{DINO}} + \mathcal{L}_{\text{iBOT}} + \lambda_k \mathcal{L}_{\text{KoLeo}},
\end{equation}
where $\lambda_k = 0.1$. The DINO and iBOT objectives provide complementary global and local supervision, while KoLeo prevents representational collapse by maintaining uniform feature dispersion. 
The teacher parameters are updated via exponential moving average:
\begin{equation}
\boldsymbol{\theta}_t \leftarrow m\,\boldsymbol{\theta}_t + (1{-}m)\,\boldsymbol{\theta}_s,
\end{equation}
where $m$ follows a cosine schedule from 0.994 to 0.999.

\subsection{Pretraining Configuration}\label{sec:pretraining_config}
We use AdamW ($\beta_1{=}0.9$, $\beta_2{=}0.999$) for pretraining with a cosine learning-rate schedule. For our base learning rate, we took $1{\times}10^{-4}$, linearly scaled with the effective batch size, and we set the number of warm-up epochs to 10 and the minimum learning rate to $10^{-6}$. More importantly, we apply the full learning rate (no scaling) to the backbone, enabling an honest reshaping of features from our ImageNet-12K initialization to the agricultural representation. We use a weight decay of 0.05 and a gradient clipping value of norm 3.0. Our teacher momentum for EMA is between 0.994 and 0.999 with a cosine schedule. 

We create 2 global crops, ($224{\times}224$) for each image, using agriculture-adapted augmentations of scale 0.5--1.0. We add reduced Gaussian blur ($\sigma \in [0.1, 1.0]$), to preserve the texture of diseases, include rotation ($\pm 15^\circ$), and use asymmetric color jitter (which we apply higher on the student view and lower on the teacher view, giving cleaner targets).

All pretraining has been performed on 8$\times$ NVIDIA H200 SXM GPUs (141 GB HBM3e per GPU) with distributed data-parallel bfloat16 mixed-precision training. We set our per-GPU batch size to 256, thus effectively 2048, as it is a batch of 8x256.

\subsection{Task-Specific Fine-Tuning}\label{sec:finetuning}
For downstream tasks, we attach a classification head comprising(LayerNorm, Dropout, and a Linear layer) and fine-tuning the model in two phases: initially freezing the feature extractor and subsequently unfreezing it for end-to-end optimization. Fine-tuning uses AdamW with a cosine schedule, per-GPU batch size of 256, label smoothing of 0.1, MixUp ($\alpha{=}0.2$), CutMix ($\alpha{=}1.0$), and early stopping on validation accuracy (patience 10--15 epochs). For class-imbalanced datasets, we employ a distributed weighted sampler and class-weighted cross-entropy loss.

For the \textit{linear probe} evaluation, the backbone is entirely frozen and only the linear classification head is trained, measuring the quality of pretrained representations without any task-specific adaptation.

%% file: 3_results.tex
\begin{table}[!htbp]
\centering
\footnotesize
\caption{A100 vs ImageNet-12K Baseline across evaluation protocols (Test Accuracy \%)}
\label{tab:main_results}
\resizebox{\columnwidth}{!}{
\begin{tabular}{>{\centering\arraybackslash}p{1.8cm} l >{\centering\arraybackslash}p{1.3cm} >{\centering\arraybackslash}p{1.3cm}}
\toprule
\shortstack{\textbf{Evaluation}\\\textbf{Protocol}}  & \textbf{Method} & \shortstack{\textbf{Crop}\\\textbf{Identification}} &
\shortstack{\textbf{Disease}\\\textbf{Detection}} \\
(data fraction) & & (32 cls) & (44 cls) \\
\midrule
\multirow{2}{*}{Fine-tune (100\%)} 
 & \textbf{A100} & \textbf{99.90} & \textbf{99.02} \\
 & ImageNet-12K & 99.90 & 99.00 \\
\midrule
\multirow{2}{*}{Fine-tune (10\%)} 
 & \textbf{A100} & \textbf{99.72} & \textbf{95.94} \\
 & ImageNet-12K & 98.69 & 94.82 \\
\midrule
\multirow{2}{*}{Fine-tune (1\%)} 
 & \textbf{A100} & \textbf{98.01} & \textbf{86.10} \\
 & ImageNet-12K & 95.20 & 83.90 \\
\midrule
\multirow{2}{*}{Linear Probe} 
 & \textbf{A100} & \textbf{95.37} & \textbf{96.76} \\
 & ImageNet-12K & 92.81 & 95.79 \\
\midrule
\multicolumn{2}{l}{\textit{A100 advantage}} & & \\
\multicolumn{2}{l}{\scriptsize ~~Fine-tune (1\%)} & \textit{+2.81} & \textit{+2.20} \\
\multicolumn{2}{l}{\scriptsize ~~Linear Probe} & \textit{+2.56} & \textit{+0.97} \\
\bottomrule
\end{tabular}}
\end{table}

\begin{figure}[h]
    \centering
    \includegraphics[width=\columnwidth]{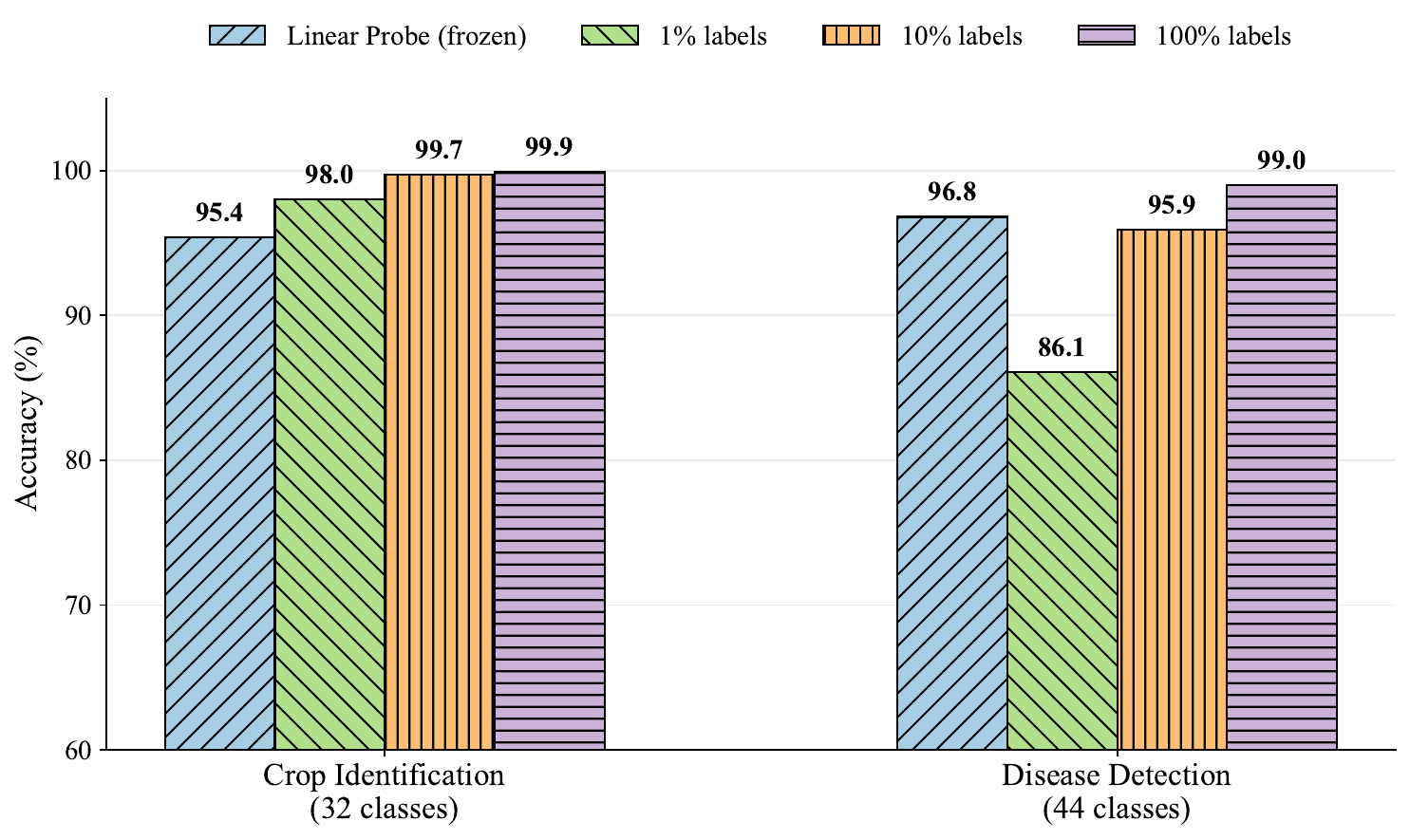}
    \caption{A100 Performance Across Data Regimes. Label efficiency of A100 under limited supervision (1\% and 10\% labels) compared to the fully supervised (100\%) and frozen-backbone (linear probe) setups on crop identification and disease detection.}
    \label{fig:label_efficiency}
\end{figure}

\begin{table}[!htbp]
\centering
\footnotesize
\caption{External Benchmark Performance (Test Accuracy \%)}
\label{tab:multidataset}

\begin{tabular}{lccc}
\toprule
\textbf{Model} & \textbf{PlantVillage} & \textbf{PlantDoc} & \textbf{PlantWild} \\
 & (38 cls) & (28 cls) & (89 cls) \\
\midrule
\textbf{A100 (Ours)} & \textbf{99.96} & \textbf{64.89} & \textbf{76.96} \\
Hybrid Baseline (IN-12K)$^\ddagger$ & 99.96 & 60.59 & 75.75 \\
ResNet-101 (IN)$^\dagger$ & 99.71 & 57.63 & 73.88 \\
Swin-T (IN)$^\dagger$ & 99.73 & 59.75 & 76.32 \\
CvT-13 (IN)$^\dagger$ & 99.73 & 63.56 & 77.31 \\
MobViT-S (IN)$^\dagger$ & 99.34 & 50.00 & 70.40 \\
\bottomrule
\multicolumn{4}{l}{\scriptsize $^\dagger$ ImageNet-pretrained baselines from prior work.}\\
\multicolumn{4}{l}{\scriptsize $^\ddagger$ Same architecture as A100, ImageNet-12K init (no SADG).}\\
\multicolumn{4}{l}{\scriptsize All benchmarks excluded from A100 pretraining corpus.}
\end{tabular}
\end{table}


\section{Results and Analysis}\label{headings}

We have evaluated A100 on three downstream tasks and three external benchmarks. We report top-1 accuracy on held-out test sets, and additionally macro-averaged F1 for the comparison with agricultural foundation models, since the crop-identification classes are imbalanced (251 to 15,455 test images per class).
We investigated the utility of A100 as a released foundation model by further analyzing its label efficiency and linear probe performance. Further, for all the downstream tasks, we have used the same pre-trained A100 checkpoint.
 
\subsection{Comparison with State-of-the-art}
Table~\ref{tab:main_results} shows a detailed comparison of the A100 and a supervised baseline (ImageNet-12K) on four different evaluation protocols, all based on the same architecture, but with varying degrees of supervision.

With full fine-tuning (100\% labels), both initializations converge to near-identical accuracy: 99.90\% (crop) and 99.02\% vs 99.00\% (disease). This is in line with the conclusion from transfer learning studies that the choice of the pretraining has little effect when there is ample labeled data and full backbone adaptation.

It is clear that A100 is more efficient than ImageNet-12K as the amount of labeled data decreases. At 10\% labels, A100 outperforms ImageNet by +1.03\% on crop identification (99.72\% vs 98.69\%) and +1.12\% on disease detection (95.94\% vs 94.82\%). At 1\% labels- the most practically relevant regime for agricultural deployments- the gap widens further to +2.81\% on crop (98.01\% vs 95.20\%) and +2.20\% on disease (86.10\% vs 83.90\%). The linear probe evaluation, in which the backbone is not unlocked at any point, shows substantial gains: Crop Identification +2.56\% and Disease Detection +0.97\%.

This is a steady increase, as the representations learned during A100 pretraining are directly applicable to the visual patterns encountered in agriculture, with the highest gains observed in minimal contexts such as real agricultural deployments.

Beyond this same-architecture control, we compare A100 directly with publicly released agricultural foundation models in Section~\ref{sec:agri_fm}.

\begin{table}[!htbp]
\centering
\footnotesize
\caption{Comparison with publicly released agricultural foundation models on crop identification (32 classes, 114,260 test images) with 1\% of the training labels (9,136 images).}
\label{tab:agri_fm}
\setlength{\tabcolsep}{2.5pt}
\resizebox{\columnwidth}{!}{
\begin{tabular}{llccc}
\toprule
\textbf{Model} & \textbf{Pretraining data} & \textbf{Params} & \textbf{Acc. (\%)} & \textbf{Macro-F1 (\%)} \\
\midrule
\textbf{A100 (Ours)} & 3M agri., 100+ crops & 73M & \textbf{98.01} & \textbf{97.10 $\pm$ 0.12}$^\S$ \\
ImageNet-12K (same arch.) & ImageNet-12K (sup.) & 73M & 95.20$^\ast$ & 93.26 $\pm$ 0.17 \\
FoMo4Wheat~\cite{han2025fomo4wheat} ViT-B/14 & 2.5M wheat & 87M & 83.86 $\pm$ 0.28 & 79.70 $\pm$ 0.36 \\
LeafVision~\cite{han2026leafvision} ResNet-50 & 540K leaf & 24M & 79.25 $\pm$ 0.79 & 76.18 $\pm$ 0.68 \\
LeafVision~\cite{han2026leafvision} ViT-B/16 & 540K leaf & 86M & 77.83 $\pm$ 0.38 & 74.77 $\pm$ 0.35 \\
\bottomrule
\multicolumn{5}{l}{\scriptsize $^\ast$ Re-run over three seeds with the same pipeline: 95.43 $\pm$ 0.07.}\\
\multicolumn{5}{l}{\scriptsize $^\S$ Mean $\pm$ std over repeated runs.}
\end{tabular}}
\end{table}

\subsection{Comparison with Agricultural Foundation Models}\label{sec:agri_fm}
We compare A100 with the agricultural foundation models whose pretrained weights are publicly available: FoMo4Wheat~\cite{han2025fomo4wheat} (ViT-B/14, pretrained on 2.5M wheat images) and LeafVision~\cite{han2026leafvision} (DINO-pretrained ResNet-50 and ViT-B/16 on 540K leaf images). Each model is fine-tuned end-to-end on the same 1\% crop-identification subsets with the same classification head (LayerNorm, Dropout, Linear) and the same recipe as A100 (Section~\ref{sec:finetuning}), using its native input normalization; for ViT backbones the feature is the concatenation of the [CLS] token and the mean patch token. Each model is evaluated on the full 114,260-image test set for each of three subset seeds.

Table~\ref{tab:agri_fm} shows that A100 leads the strongest public agricultural model, FoMo4Wheat, by \textbf{14.2 points}, and both LeafVision models by about 19--20 points. Both existing agricultural models also fall below the generic ImageNet-12K initialization in this setting, and the macro-F1 ranking is identical. These models are strong in their own domains, wheat canopy and organ imagery for FoMo4Wheat and leaf-level disease imagery for LeafVision, but their representations do not transfer to multi-crop in-field identification. This is precisely the gap that A100's 100+-crop field corpus targets. Agri-FM+~\cite{alnahian2025agrifm} could not be included, because its pretrained weights have not been publicly released and it reports only detection and segmentation results.

\subsection{Label Efficiency}
Figure \ref{fig:label_efficiency} shows the absolute performance of A100 in the limited supervision scenario that is most relevant for agricultural applications, where expert annotation is costly.

The most striking result is crop classification: A100 is able to outperform the ImageNet-12K baseline (+2.81\%) with just a 1\% of labeled data, namely (${\sim}$9{,}100 images) across 32 classes. This is a modest, yet achievable, budget for agricultural extension service labels: ${\sim}$284 labels per crop. The accuracy is 99.72\% at 10\% labels ( ${\sim}$91K images), which is better than ImageNet (+1.03\%) and close to the fully supervised maximum.

To check that the low-label gain is not sensitive to the sampled subset, we re-ran the 1\% ImageNet-12K baseline over three independent per-class subset seeds (42, 123, 2026), each evaluated on the full test set, obtaining $95.43 \pm 0.07\%$ (95.35 / 95.49 / 95.45), consistent with the 95.20\% in Table~\ref{tab:main_results}. The $+2.81$-point gap between A100 and ImageNet-12K is therefore about $40\times$ the seed standard deviation. The public agricultural models in Table~\ref{tab:agri_fm} likewise vary by less than 1 point across the same seeds.

The benefit of domain-specific pretraining is even more clear at low data for disease detection. With only 1\% labels (418 images total, ${\sim}$9.5 per class), A100 achieves 86.10\% compared to ImageNet's 83.90\% (+2.20\%). At 10\% labels (4{,}200 images across 44 disease categories), A100 yields 95.94\% vs ImageNet's 94.82\% (+1.12\%). The improvements demonstrate that SADG pre-training has learned discriminative patterns of diseases that are easier to trigger without annotation for the task.

\subsection{External Benchmark Evaluation}
To assess generalization, we evaluate A100 on three widely-used plant disease benchmarks (Table~\ref{tab:multidataset}). Importantly, \textit{all three datasets are completely omitted from the pretraining corpus}, guaranteeing unbiased evaluation.

\textbf{PlantVillage} (38 classes, 54K images): A100's accuracy is the highest reported on this dataset without data contamination:\textbf{99.96\%}. It outperforms all ImageNet-pretrained baselines, including CvT-13 (99.73\%), thereby demonstrating that domain-specific pretraining yields better features for controlled plant disease imagery. 

\textbf{PlantDoc} (28 classes, 2.3K images): A more difficult benchmark which includes some noisy images with web-scraped images in natural conditions. A100 achieves 64.89\%, substantially outperforming the ImageNet-12K baseline (60.59\%, +4.30\%) and all other baselines including CvT-13 (63.56\%, +1.33\%). This is the strongest evidence that domain-specific pre-training yields better features for hard real-world plant images with only a few training images.

\textbf{PlantWild} (89 classes, 18.5K images): Wild images with 89 fine-grained classes of diseases. A100 achieves 76.96\%, outperforming the ImageNet-12K baseline (75.75\%, +1.21\%) and approaching CvT-13 (77.31\%) and substantially above MobViT-S (70.40\%).

It also shows that the external benchmark results are best when using A100's agricultural pretraining on challenging data-scarce benchmarks (PlantDoc: +4.30\%) for which generic ImageNet features are not enough.

\subsection{Representation Quality: Linear Probe Analysis}
Table~\ref{tab:main_results} compares the quality of the representations produced by A100 with those produced by ImageNet-12K supervised pretraining, using a frozen-backbone linear probe, as is standard in the literature for assessing the quality of representations produced by a model for representation learning.

A100's agricultural representations, trained in a self-supervised manner, outperform ImageNet features on \textbf{plant-centric tasks} (+2.56\% on crop identification, +0.97\% on disease detection). This indicates that, even without labels, domain-specific pretraining on 3M agricultural images yields features more relevant to plant morphology and pathology than supervised pretraining on 12M generic ImageNet images.

The linear probe evaluation is relevant to practice. For users lacking full fine-tuning capabilities due to limited compute resources and a lack of GPUs, A100 features can be used directly with a simple linear classifier to classify crops with 95.37\% accuracy and detect diseases with 96.76\% accuracy, making it a compelling deployment option for resource-constrained agricultural settings.

\subsection{Beyond Classification: Plant Phenotyping (Leaf Counting)}

To show how A100's representations can be used beyond classification to regression-based phenotyping tasks, we evaluate on the leaf-counting task using the GroMo25 dataset~\cite{gromo2025} of multiview time-series plant image data for plant growth modeling (ACM Multimedia 2025 Grand Challenge). The dataset consists of high-resolution images of 4 crop species (Wheat, Okra, Radish, Mustard), taken in controlled conditions and along with the image, annotated with a count of leaves.

\begin{figure}[!t]
    \centering
    \includegraphics[width=\columnwidth]{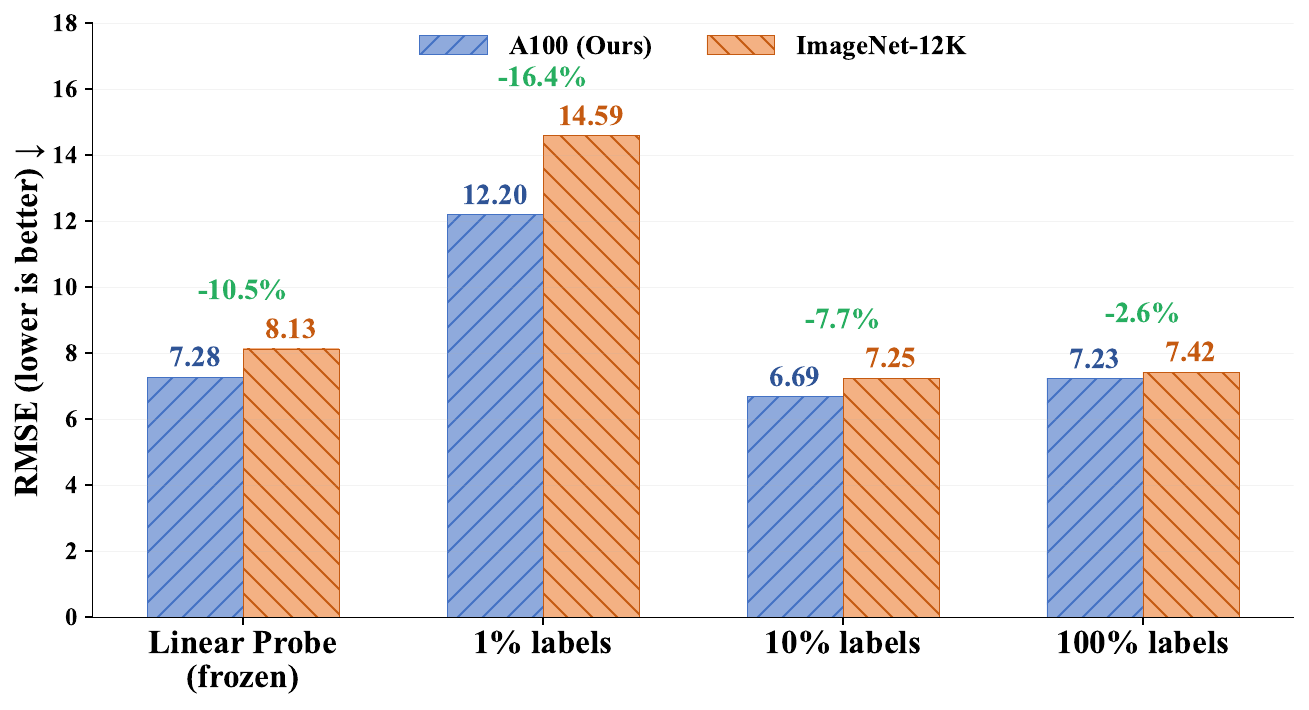}\\[-2pt]
    \includegraphics[width=\columnwidth]{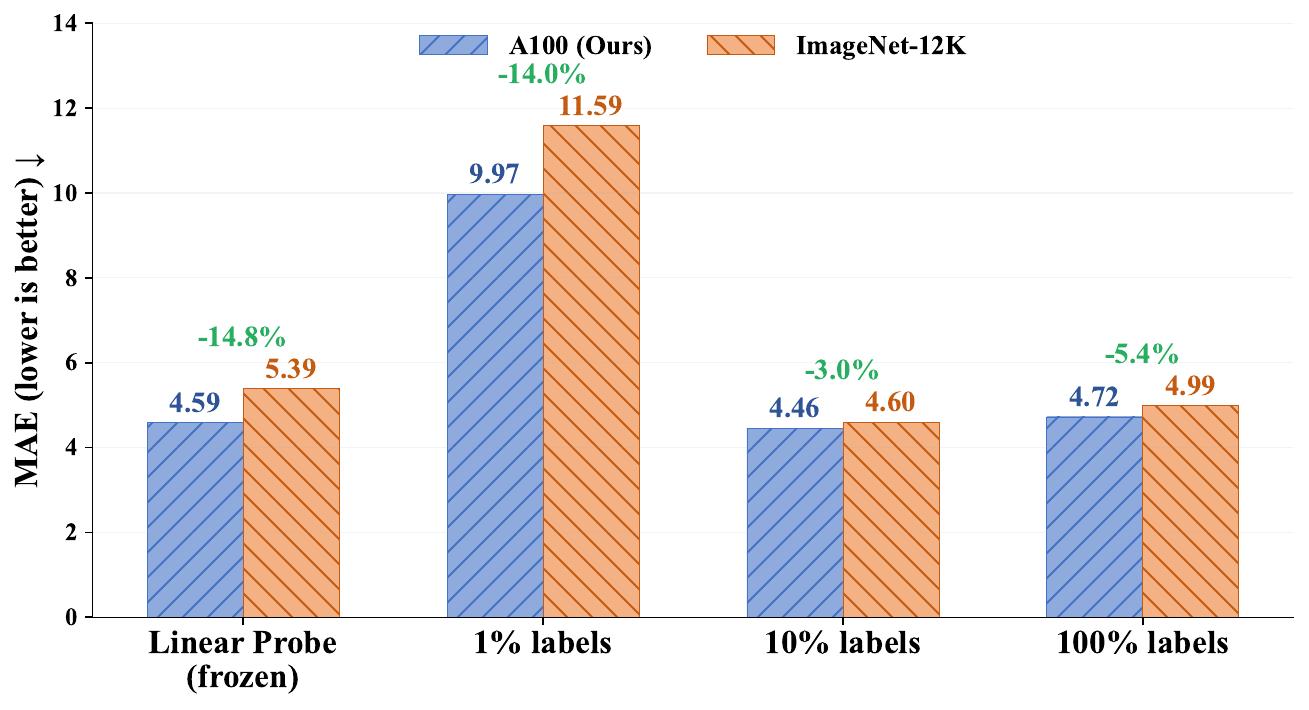}
    \caption{Plant phenotyping (leaf counting): A100 vs ImageNet-12K. (Top) RMSE and (Bottom) MAE. A100 consistently reduces prediction error across all protocols, with the largest improvement at 1\% labels (-16.4\% RMSE, -14.0\% MAE).}
    \label{fig:leaf_counting}
\end{figure}

Fig.~\ref{fig:leaf_counting} shows results for leaf counting, a task distinct from classification which involves estimating a continuous value for the number of leaves in a plant from a single image of that plant at all growth stages (2--40 leaves). On \textit{all} evaluation protocols A100 surpasses ImageNet-12K:

\begin{itemize}
    \item \textbf{Linear probe:} A100 achieves RMSE 7.28 vs ImageNet's 8.13 (\textbf{10.4\% lower error}), demonstrating that A100's frozen features encode plant structural information useful for counting without any adaptation.
    \item \textbf{1\% labels:} The advantage grows to \textbf{16.4\%} (RMSE 12.20 vs 14.59), confirming that A100's features are especially valuable under extreme label scarcity.
    \item \textbf{Full fine-tuning:} A100 still maintains a measurable advantage (7.23 vs 7.42), unlike classification where both initializations converge.
\end{itemize}

This result is significant for two reasons. First, it shows that A100 not only supports classification but also \textbf{regression-based phenotyping}, which involves understanding of plant architecture and growth stage, not just categorical identity. Second, while images captured by the GroMo25~\cite{gromo2025} are taken under controlled indoor conditions (white background, rotating platform), this differs markedly from the field-image corpus used for pretraining the A100, yet the A100 still outperforms ImageNet. This is a real example of cross-domain transfer of agricultural visual knowledge.

\subsection{Discussion}

\textbf{Domain-specific pretraining advantage.} The best results to come out of A100's use seem to be from the PlantDoc benchmark (+4.30\% over identical architecture with ImageNet-12K init), and from the linear probe comparison (+2.56\% on crop identification). These improvements come right where they are needed, in scenarios where there is limited training data and backbone adaptation isn't possible.

\textbf{Source of the gains.} Two comparisons separate the contributing factors. Holding the architecture fixed, A100 versus the identical CoAtNet-2 initialized from ImageNet-12K isolates what self-supervised pretraining on the 3M agricultural images adds: +2.81 points at 1\% labels, +2.56 in the linear probe and +4.30 on PlantDoc. Varying the pretraining corpus instead, the public agricultural models trained with established self-supervised methods on narrower corpora (Table~\ref{tab:agri_fm}) trail A100 by 14--20 points at 1\% labels, which indicates that the breadth of the corpus (100+ crops, in-field imagery), and not agricultural pretraining alone, is the dominant factor for multi-crop field recognition.

\textbf{Task generalization.} A100 can be applied beyond the classification. We quantitatively assess the leaf-counting ability, achieving a 10.4\% decrease in Root Mean Square Error (RMSE) with linear probing, demonstrating that the model's learned representations capture intrinsic plant morphological features suitable for continuous regression. The results are consistent with the main principle of foundation models: a single pretrained feature extractor can be easily fine-tuned for a wide range of visual applications in agriculture.

\textbf{Practical deployment.} The label efficiency results highlight the main advantage of A100 for release: agricultural actors can achieve production-grade accuracy ($>$98\%) with less than 300 labeled images per class, which is an important factor in data-limited regions. When paired with the capabilities of a linear probe (no fine-tuning required), A100 reduces the cost of bringing agricultural AI to market.

\textbf{Limitations.} The PlantDoc results (64.89\%) suggest issues with small, noisy training sets under free conditions, although A100 still outperforms all baselines. The current study evaluates a single backbone architecture (CoAtNet-2); extending the pretraining framework to other architectures (ViT, Swin) remains future work, as does a component-wise ablation of the three SSL objectives (DINO, iBOT, KoLeo) on our corpus. The corpus was deduplicated with file-name and file-size heuristics; perceptual-hash near-duplicate detection within the corpus and against all downstream and benchmark images would provide a stronger guarantee and is left to future work.